\documentclass[conference]{IEEEtran}
\IEEEoverridecommandlockouts
\usepackage{booktabs}
\usepackage{cite}
\usepackage{amsmath,amssymb,amsfonts}
\usepackage{algorithmic}
\usepackage{graphicx}
\usepackage{textcomp}
\usepackage{xcolor}
\usepackage{orcidlink}
\begin{document}

\title{LLM-Driven Ontology Construction for Enterprise~Knowledge~Graphs}

\author{
\IEEEauthorblockN{Abdulsobur Oyewale}
\IEEEauthorblockA{\textit{Liber AI Research} \\ \textit{London, United Kingdom} \\
\href{mailto:abdulsobur@liberai.org}{abdulsobur@liberai.org}}
\and
\IEEEauthorblockN{Tommaso Soru}
\IEEEauthorblockA{\textit{Liber AI Research} \\ \textit{London, United Kingdom} \\
\href{mailto:tom@liberai.org}{tom@liberai.org} \orcidlink{0000-0002-1276-2366}}
}

\maketitle

\begin{abstract}
Enterprise Knowledge Graphs have become essential for unifying heterogeneous data and enforcing semantic governance.
However, the construction of their underlying ontologies remains a resource-intensive, manual process that relies heavily on domain expertise.
This paper introduces OntoEKG, a LLM-driven pipeline designed to accelerate the generation of domain-specific ontologies from unstructured enterprise data.
Our approach decomposes the modelling task into two distinct phases: an extraction module that identifies core classes and properties, and an entailment module that logically structures these elements into a hierarchy before serialising them into standard RDF.
Addressing the significant lack of comprehensive benchmarks for end-to-end ontology construction, we adopt a new evaluation dataset derived from documents across the Data, Finance, and Logistics sectors.
Experimental results highlight both the potential and the challenges of this approach, achieving a fuzzy-match F1-score of 0.724 in the Data domain while revealing limitations in scope definition and hierarchical reasoning.
\end{abstract}

\begin{IEEEkeywords}
artificial intelligence, large language models, semantic models, ontology construction, RDF, knowledge graphs
\end{IEEEkeywords}

% [x] write intro with overview
% [x] add example text + derived ontology as graph
% [x] write related work (search for llm-driven ontology construction, EKG)
% [x] make pipeline diagram
% [x] formalise the approach
% [x] perform evaluation on benchmark
% [x] add eval results in table
% [x] compute fuzzy-match eval with text embeddings
% [ ] qualitative analysis
% [ ] write abstract
% [ ] write conclusion
% [ ] publish source code

\section{Introduction}

In the last decade, enterprises have increasingly embraced semantic technologies and the Resource Description Framework (RDF) to unify heterogeneous data sources, enforce shared meaning, and enable interoperable analytics across business domains.
This shift reflects the growing recognition that enterprise data assets require explicit semantics to support governance, lineage, and downstream intelligence at scale.
Ontologies play a central role in these ecosystems by capturing conceptual structure, constraining vocabularies, and providing the backbone for enterprise knowledge graphs (EKGs).

\begin{figure}[t]
    \centering
    \includegraphics[width=0.9\linewidth]{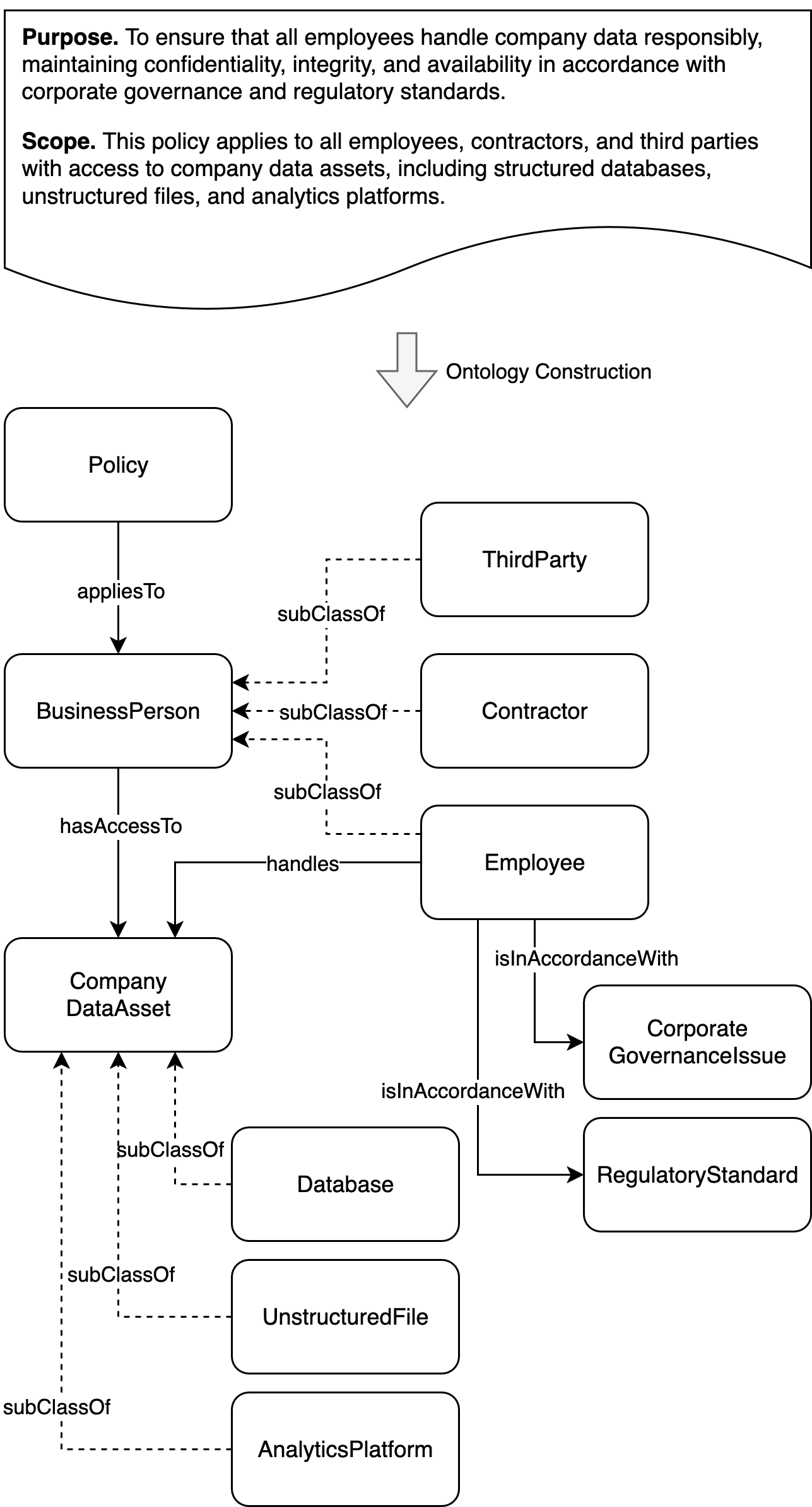}
    \caption{An example of Ontology Construction task.}
    \label{fig:ontologyconstruction}
\end{figure}

At the same time, neural models and more recently large language models (LLMs) have begun to transform data engineering and integration workflows.
Their ability to extract semantics from unstructured content, suggest schema patterns, perform question answering, and align business terminology introduces powerful new opportunities for knowledge graph (KG) construction~\cite{bian2025llm_survey,soru2017sparql}.
When embedded in transformation pipelines, LLMs offer the potential to accelerate ontology engineering tasks that previously depended on extensive human interpretation and domain expertise.

Yet ontology construction in enterprises remains largely manual, iterative, and resource-intensive.
Domain stakeholders, data architects, and semantic engineers must repeatedly negotiate conceptual boundaries, align schemas, and document modelling decisions.
This motivates the development of an AI-based copilot for ontology design --- one that collaborates with human experts while maintaining rigour, transparency, and governance compliance.
In this paper, we explore how LLM-driven workflows can support the creation and evolution of ontologies specifically for enterprise knowledge graphs, reducing modelling friction while preserving semantic quality.

This paper is organised as follows.
Related work is introduced in~\autoref{sec:related}.
We describe the approach in~\autoref{sec:approach}.
We discuss results in~\autoref{sec:evaluation}.
Finally, we conclude.

\section{Related Work} \label{sec:related}

Ontology extraction for the enterprise has been a research topic for more than two decades. One of the first works in this area, titled \textit{A Method for Semi-Automatic Ontology Acquisition from a Corporate Intranet}, proposed a comprehensive architecture and methodology that leverages corporate intranet text and semi-structured resources to automatically extract and refine a domain-specific ontology, reducing reliance on purely manual ontology engineering~\cite{kietz2000method}.

Much more recently, EOAC-LLM has been introduced as a five-step approach based on LLMs to enable the automatic generation of domain-specific event ontologies. Unlike us, the authors focus on a multi-dimensional aggregation method for semantic temporal relation~\cite{lu5775146eoac}. 

The paper \textit{Ontology Generation using Large Language Models} presents two prompting techniques, Memoryless CQbyCQ and Ontogenia. Differently from our method, these emphasise multi-dimensional evaluation including structural criteria, alongside expert assessment~\cite{lippolis2025ontology}.

Also, a recent paper, \textit{From human experts to machines: An LLM supported approach to ontology and knowledge graph construction} explores semi-automatic construction of KGs facilitated by open source LLMs by formulating competency questions, T-Box Development, KGs population and evaluation with minimal human expert involvement. Their focus was on full KG construction, and they employed an LLM as a Judge for automatic evaluation of Retrieval-Augmented Generation (RAG) generated answers and extracted concepts~\cite{kommineni2024human}.

The paper \textit{Leveraging LLM for Automated Ontology Extraction and Knowledge Graph Generation} leverages LLM through an interactive user interface. It employs the use of an iterative Chain of Thought algorithm to allow the users to iteratively refine and confirm the ontology based on their preference~\cite{abolhasani2024leveraging}.

\textit{Navigating Ontology Development with Large Language Models} investigate LLMs capability to generate OWL ontologies from ontological requirements using various prompting techniques, and explores comparing these prompting techniques across multiple state of the art model~\cite{saeedizade2024navigating}.

\section{Approach} \label{sec:approach}

This work is centred around two main contributions:
\begin{itemize}
    \item A pipeline for Ontology Construction from text which specifically targets enterprise data;
    \item A call to develop a comprehensive benchmark for evaluating Ontology Construction from text.
\end{itemize}

An example of Ontology Construction task can be seen in~\autoref{fig:ontologyconstruction}. Starting from an input text about an enterprise in the cybersecurity sector, an RDF-based ontology is built. Dashed edges identify hierarchical structure, while solid edges identify relations between classes, specifying domain and ranges.

\subsection{Formalisation}

Our Ontology Construction task can be formalised as follows.
Given an input text $T$, we infer the set of classes $C_T$ and the set of properties $P_T$ from the corpus.
Each class $c \in C_T$ is associated with a label and a description.
Each property $p \in P_T$ is associated with a label, a domain class, and a range class such that $dom(p) \in C$, $rng(p) \in C$.
Classes may also exist in a hierarchy, where $c_1 \subseteq c_2$ means all elements that belong to class $c_1$ are also in class $c_2$.

In RDF terms, each class $c$ is an instance of owl:Class and each property $p$ is an instance of owl:ObjectProperty.
In OntoEKG, datatypes are reified into their own classes as in \textit{Schema.org}\footnote{\url{https://schema.org/DataType}}.

\subsection{Pipeline}

\begin{figure}[h]
    \centering
    \includegraphics[width=1\linewidth]{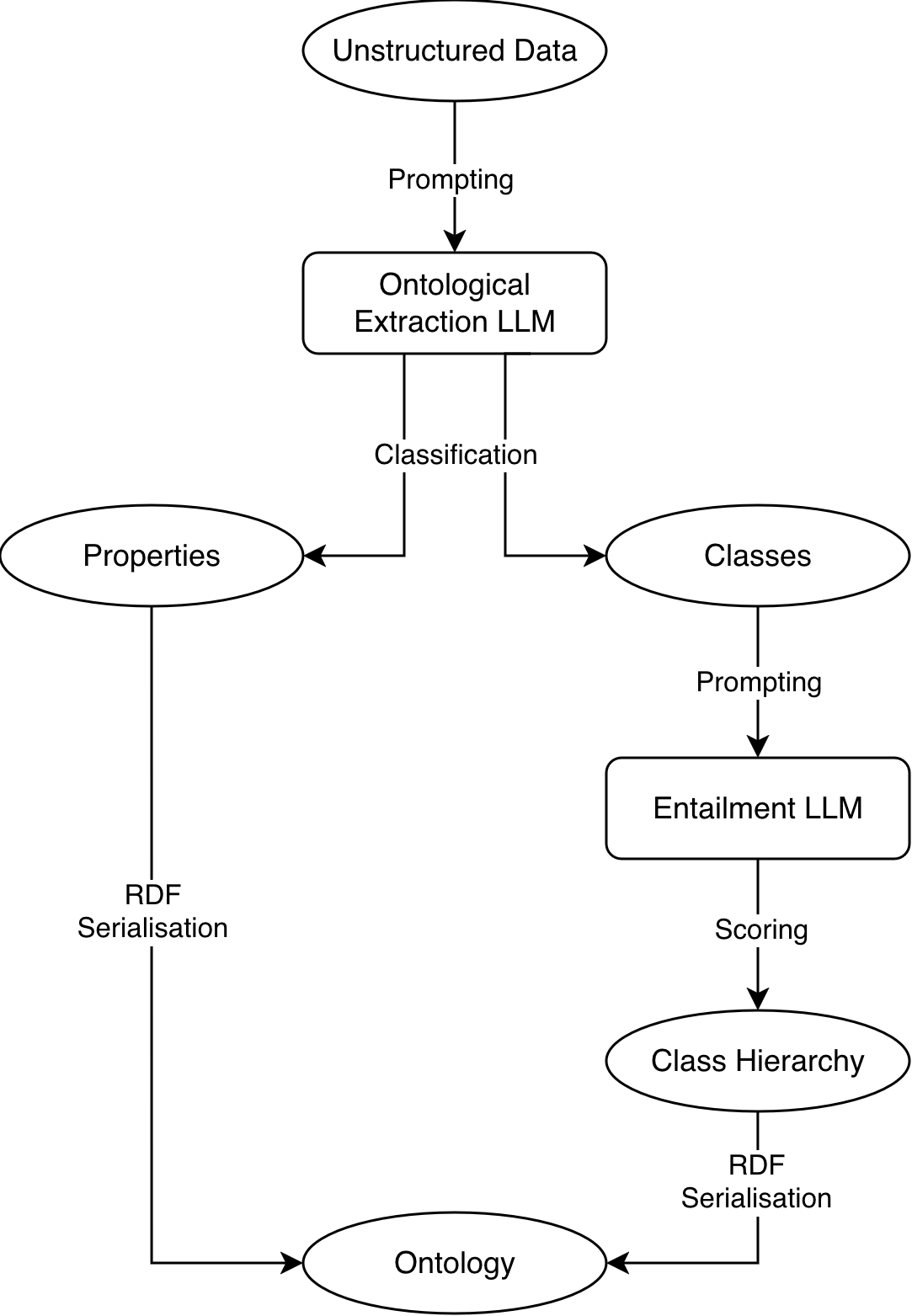}
    \caption{The full OntoEKG pipeline.}
    \label{fig:pipeline}
\end{figure}

\autoref{fig:pipeline} shows how the OntoEKG pipeline is structured.

The pipeline transforms unstructured enterprise text into a structured Ontology ready for Knowledge Graphs. It utilises a two-step LLM process: first to extract classes and properties, and second to reason about the hierarchical relationships between those classes. Below is a detailed workflow of each steps:

\begin{enumerate}
    \item \textbf{Data Ingestion}. Our pipeline process begins by feeding unstructured data into the system. To ensure the expected output is structured, we define strict data models using Pydantic. This forces our LLM to output valid JSON containing specific metadata like classes, properties, description, domain, range.

    \item \textbf{Ontological Element Extraction}. The raw unstructured data is processed by the Ontology Extraction LLM. The pipeline uses a specialised system prompt to identify two core elements: Classes (e.g., types of entities like ``Employee'' or ``Vehicle'') and Properties (relationships like ``operates'' or ``hasAccessTo''). We strictly crafted our prompts to focus solely on the schema we provided it with.

    \item \textbf{Hierarchy Construction with Entailment}. For the list of classes outputted by our Ontological Element Extraction module, this step organises this list of classes into a logical taxonomy. An Entailment LLM iteratively analyses the extracted classes to determine inheritance relationships (e.g., verifying if ``Apple'' is a subclass of ``Fruit''). It uses logical reasoning to validate these relationships based on the class descriptions.

    \item \textbf{RDF Serialisation}. Finally, the system merges the extracted properties and the constructed hierarchy into a formal graph. Using the rdflib library, it converts the data into structured RDF triples (i.e., using owl:Class and owl:ObjectProperty) and saves the result to a Turtle file, creating a machine-readable ontology.
\end{enumerate}

\section{Evaluation} \label{sec:evaluation}

\subsection{Benchmarks}

As mentioned in~\autoref{sec:approach}, a primary objective of this paper is to issue a call to action for the research community.
Specifically, to the best of our knowledge, there is a lack of comprehensive benchmarks for evaluating Ontology Construction from text.
According to our findings, previous works either have failed to address the task in its entirety or do not meet the required quality standards.

OntoURL is a comprehensive benchmark for evaluating LLMs' capabilities in handling ontologies across three dimensions: understanding, reasoning, and learning.
The second and third task of the learning category target class hierarchy construction and property relation construction, respectively~\cite{zhang2025ontourl}.
However, these tasks expect a semi-structured input which includes the list of classes and properties.
Therefore, OntoEKG and other approaches that extract knowledge from purely unstructured data cannot be evaluated using this benchmark.

Benchmarks such as Text2KGBench and OSKGC emphasise instance-level extraction within an existing framework, treating the ontology as a constraint rather than the end product~\cite{mihindukulasooriya2023text2kgbench,wang2025oskgc}.

The LLMs4OL initiative hosts a challenge open to LLM-based approaches for Ontology Learning divided into four tasks.
Task A (Text2Onto) targets term and type extraction from text; here, the expected output is a list of potential terms for an ontology, while Task C targets Taxonomy Discovery~\cite{giglou2025llms4ol}.
Unfortunately, the tasks are not arranged in series, which makes it impossible to evaluate a full Ontology Construction pipeline on them.
Moreover, Task C mixes class terms with individuals, making no distinction between T-Box and A-Box.

Given the findings above, we opted for the creation of our own dataset, which consists of three use cases --- excerpts from internal enterprise policy text in the sectors of data, finance, and logistics.
Source code and data can be found at the OntoEKG repository on Github\footnote{\url{https://github.com/LiberAI/OntoEKG}}.

\subsection{Experiments}

% eval setup, LLM(s) used, runtime...
We ran our experiments on a cloud machine provided by Google Colab\footnote{\url{https://colab.research.google.com}}.
For the Ontological Extraction step, we chose Google Gemini 3 Flash (preview)~\cite{Google_Gemini_2025}; for the Entailment step, we chose Anthropic Claude 4.5 Opus~\cite{Anthropic_Claude_4.5_Opus_2025}.

\begin{table}[h]
    \centering
    \caption{Ontology Construction performance, exact match.}
    \label{tab:exactmatch}
    \begin{tabular}{lccc}
        \toprule
        \textbf{Use case} & \textbf{Precision} & \textbf{Recall} & \textbf{F1} \\
        \midrule
        \textbf{Data} & 0.083 & 0.133 & 0.102 \\
        \textbf{Finance} & 0.000 & 0.000 & 0.000 \\
        \textbf{Logistics} & 0.040 & 0.062 & 0.048 \\
        \bottomrule
    \end{tabular}
\end{table}

\begin{table}[h]
    \centering
    \caption{Ontology Construction performance, fuzzy match.}
    \label{tab:fuzzymatch}
    \begin{tabular}{lccc}
        \toprule
        \textbf{Use case} & \textbf{Precision} & \textbf{Recall} & \textbf{F1} \\
        \midrule
        \textbf{Data} & 0.656 & 0.807 & 0.724 \\
        \textbf{Finance} & 0.095 & 0.166 & 0.121 \\
        \textbf{Logistics} & 0.366 & 0.523 & 0.431 \\
        \bottomrule
    \end{tabular}
\end{table}

\begin{figure*}[t]
    \centering
    \includegraphics[width=0.9\linewidth]{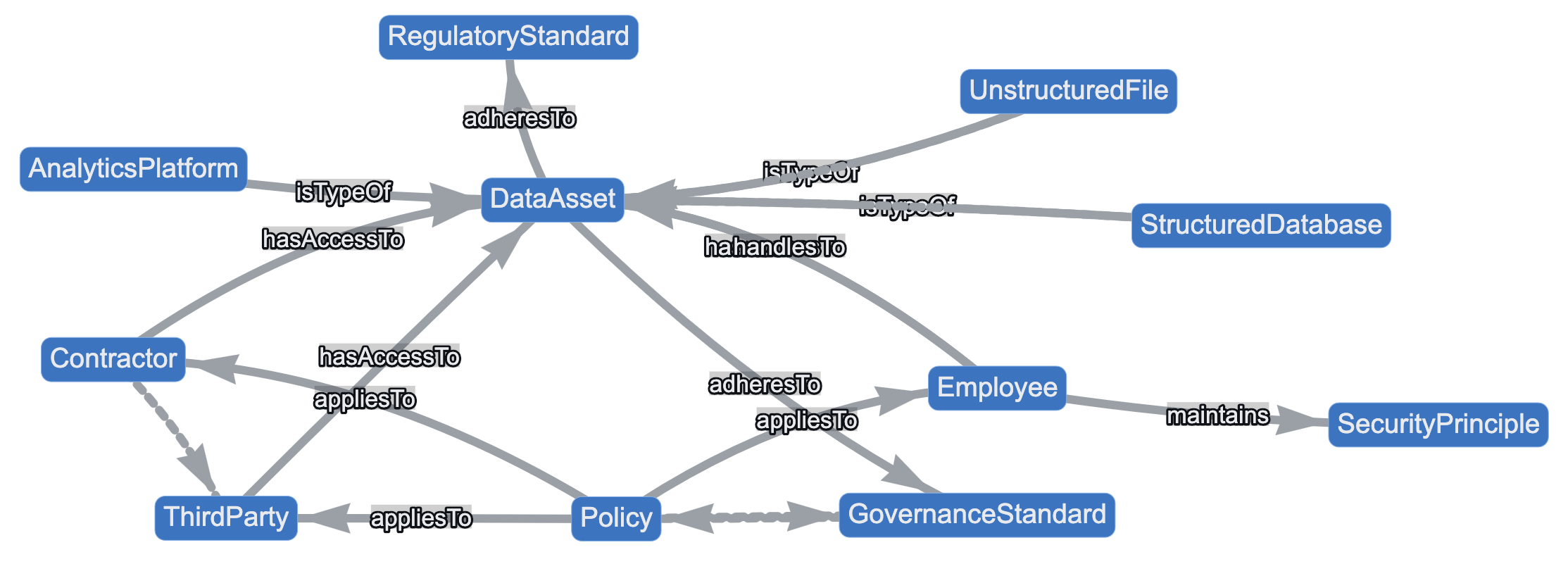}
    \caption{The generated ontology for the Data use case.}
    \label{fig:output}
\end{figure*}

Results can be seen in~\autoref{tab:exactmatch} and~\autoref{tab:fuzzymatch}.
\autoref{tab:exactmatch} shows OntoEKG's performance considering only triples that match exactly.
In~\autoref{tab:fuzzymatch}, we adopted embedding-based fuzzy matching to align predicted triples with their gold standard; we set a similarity threshold of $(0.94, 0.94, 0.95)$ for the three use cases.
The best performances were reached in the Data use case, where we had an F1-score of 0.724 in the fuzzy setting.
The Finance use case was the most challenging with only 0.121 F1-score; this is likely due to different interpretations of the input text, specifically choosing which terms should be included in the ontology and which ones are out of context.

An example generation can be seen in~\autoref{fig:output}, processed from the text in~\autoref{fig:ontologyconstruction}.
Here, we can see two issues arising from the graph: ``Policy'' and ``GovernanceStandard'' were declared one the subclass of the other, implying an equivalence; also, a ``isTypeOf'' property was introduced, which remains ambiguous in RDF terms between rdf:subClassOf and rdf:type.

We tested the Entailment task with different LLMs, including Gemini 2.5 Flash, 2.5 Pro, 3 Flash (preview), and Claude 4.5 Sonnet.
We discarded Gemini 2.5 Pro for being inadequate in terms of efficiency.
The other models did not meet the expectations during the development phase.
These preliminary results confirm the need for a dedicated benchmark which would enable a more appropriate evaluation.

Despite the promising outcomes, we identified several limitations in our approach:
\begin{itemize}
    \item Determining the optimal scope of a model is difficult for an LLM to manage autonomously. To ensure accuracy, the model requires explicit signalling to define the boundaries of relevant classes and properties.
    \item Sometimes, the LLM tends to propose individuals instead of classes. Again, this is due to the fact no explicit requirements were declared in terms of a target level of abstraction.
    \item During the Entailment phase, we have seen LLMs confuse the directionality of the hierarchy relations and adopt loose definitions of subsumption which affected the logical consistency of the RDF model.
\end{itemize}

\section{Conclusion} \label{sec:conclusion}

In this paper, we have introduced OntoEKG, an LLM-driven approach to Ontology Construction for enterprise knowledge graphs.
The initial results suggest that a tedious and resource-intensive task such as semantic modelling can be supported by automated techniques.
We have also underscored the need for a comprehensive benchmark for Ontology Construction from unstructured data.

Future work will include the realisation of an end-to-end method for the translation of text into RDF-based semantic models.
We will integrate the possibility to handle named individuals and extract entity metadata directly from text, e.g. keeping information about provenance.
Furthermore, we expect to enable progressive construction of enterprise ontologies by feeding the existing proposed model to OntoEKG itself together with the input text, so that the model stays consistent across different source documents.
We plan to engage the research community to collaboratively develop a comprehensive benchmark for end-to-end Ontology Construction.

\bibliographystyle{plain}
\bibliography{references}

\end{document}